# CONSTRAINED BUNDLE ADJUSTMENT FOR STRUCTURE FROM MOTION USING UNCALIBRATED MULTI-CAMERA SYSTEMS

D. Huang[1,3,†], M. Elhashash[1,3,†], R. Qin[1,2,3,4,*]

[1] Geospatial Data Analytics Lab, The Ohio State University, Columbus, USA
[2] Department of Civil, Environmental and Geodetic Engineering, The Ohio State University, Columbus, USA
[3] Department of Electrical and Computer Engineering, The Ohio State University, Columbus, USA
[4] Translational Data Analytics Institute, The Ohio State University, Columbus, USA
Email: <huang.3918><elhashash.3><qin.324>@osu.edu

**Commission II, WG II/1**

**ABSTRACT:**

Structure from motion using uncalibrated multi-camera systems is a challenging task. This paper proposes a bundle adjustment solution that implements a baseline constraint respecting that these cameras are static to each other. We assume these cameras are mounted on a mobile platform, uncalibrated, and coarsely synchronized. To this end, we propose the baseline constraint that is formulated for the scenario in which the cameras have overlapping views. The constraint is incorporated in the bundle adjustment solution to keep the relative motion of different cameras static. Experiments were conducted using video frames of two collocated GoPro cameras mounted on a vehicle with no system calibration. These two cameras were placed capturing overlapping contents. We performed our bundle adjustment using the proposed constraint and then produced 3D dense point clouds. Evaluations were performed by comparing these dense point clouds against LiDAR reference data. We showed that, as compared to traditional bundle adjustment, our proposed method achieved an improvement of 29.38%.

## 1. INTRODUCTION

Using a multi-camera setup for Structure from Motion (SfM) is favorable for mapping-based applications, as it can stabilize the camera network structure and at the same time extend the field of views (FoV). Existing solutions (Heng et al., 2019) mostly rely on lab calibration, which require expert knowledge and are prohibitive to any minor changes impacting the configuration of the system. Using a multi-camera system is often seen as a good practice when performing SfM or mobile mapping at the ground-level, as using single or uncoupled cameras may face challenges. The chief one of them is that each image only has neighbors among the direction of motion, unlike aerial data acquisitions that usually collect data in a grid pattern so that the images are well tied in both horizontal and vertical directions. This challenge leads to a sub-optimal camera network causing error accumulations. In the case of 3D reconstruction of monocular video frames, the accumulated errors are typically observed as trajectory drift (i.e., errors in the estimated relative orientations) and non-rigid distortions in the reconstructed point clouds. These errors are challenging even if post-registration methods are used to register the reconstructed 3D model to a reference one (e.g., aerial-based model).

This paper proposes a bundle adjustment (BA) solution for uncalibrated cameras with overlapping views by incorporating a so-called baseline constraint which builds the link between two cameras that are static to each other. We performed experiments that compared the 3D reconstruction results individually achieved by the typical SfM method and simultaneously achieved by our proposed method. As a side study, we also identified 3D reconstruction results for different facing directions of the mounted cameras. To this end, our contributions are twofold. Firstly, we proposed a BA solution for uncalibrated cameras where they captured overlapping contents. We evaluated the proposed approaches qualitatively and quantitatively using LiDAR data as ground truth. Secondly, we analyzed the accuracy of performing SfM using video frames of a monocular camera, where the camera was mounted with different viewing angles for the same testing area.

The rest of this paper is organized as follows: **Section 2** reviews related works of multi-camera solutions for 3D reconstruction; **Section 3** introduces our proposed BA solution in detail; **Section 4** presents the experimental results, comparative studies, and our analysis; **Section 5** concludes this work.

## 2. RELATED WORK

**Structure from Motion.** SfM is the process of estimating the 3D structures from a set of 2D images. A typical SfM pipeline starts with feature extraction from the images, which can be either in a traditional way such as SIFT (Lowe, 2004), SURF (Bay et al., 2006), and ORB (Rublee et al., 2011), or using learned features (Agarwal et al., 2010). Feature matching is then performed to find the correspondences between images based on the feature descriptors, and the outliers are removed by geometric verification using the RANSAC scheme (Hartley and Zisserman, 2004). After estimating the camera poses from the filtered matching points, the depth of the points can be computed via triangulation. BA (Triggs et al., 1999) is used to refine the 3D points, camera poses, and camera parameters by minimizing the reprojection errors.

**Multi-camera solutions.** The multi-camera system for 3D reconstruction effectively increases the range of 3D scene information compared to the single-camera system due to the extended FoV of the cameras. Multi-camera schemes are mainly used in 3D applications such as autonomous vehicles, aerial

---

vehicle and mobile robots. The work introduced in (Heng et al., 2015b) utilized multiple stereo cameras with an inertial measurement unit (IMU) in a multi-sensor system. It proposed BA that refined the camera extrinsics, but required pre-calibrated stereo cameras. Their later work (Heng et al., 2019) used multiple cameras to facilitate localization and 3D scene perception for self-driving in urban and rural environments. However, it still required accurate calibration of the multi-camera system as a prerequisite using a fiducial target. Another study (Cavegn et al., 2018) also applied relative orientation constraints among cameras in their mobile mapping applications, but the relative orientation was also pre-calibrated. (Esquivel and Koch, 2013) used the rigidity constraints between the cameras without overlap to perform the calibration, but the experiments were done on small indoor datasets. A recent study (Maset et al., 2021) investigated adding relative orientation constraints without pre-calibration (Maset et al., 2020; Pierrot-Deseilligny et al., 2014; Schonberger and Frahm, 2016) for multi-camera systems with overlap and showed that the added constraints improved the accuracy compared to traditional BA. Other approaches (Häne et al., 2017; Heng et al., 2015a) did not require a fiducial target for calibration, but instead required a survey of the area to build a map for the environment before performing calibration, and these approaches only worked in compact environments such as indoor parking garages.

## 3. METHODOLOGY

In this paper, we considered the following case for uncalibrated multi-camera system: the system consists of two cameras capturing images that have overlaps with each other at time $i$ (which makes these two cameras an image pair), while the images captured at time $i$ have overlaps with the adjacent images in both frame sequences. By assuming such a multi-camera system, our approach built a weak constraint to fix the relative orientation of these two cameras in the BA. To start, our approach followed the typical SfM process, which began by extracting features from the images and performing feature matching to identify tie-points/matches. We used SIFT (Lowe, 2004) as a feature extractor and descriptor. We used randomized kd-trees (Muja and Lowe, 2014) for feature matching, where we set the number of trees to 8 and the number of searches per tree to 64. Since our data were sequential video frames, we matched each image to 20 adjacent images in the forward and backward directions in both sequences. Using the estimated tie-points, we incrementally estimated the exterior orientation of each camera while minimizing the errors through BA, in which our baseline constraint was implemented. Finally, we used the generated sparse reconstruction to produce dense point clouds using OpenMVS open-source library (Cernea, 2020).

We briefly introduced traditional BA in **Section 3.1** as the main minimization framework for the 3D reconstruction problem. Then we discussed the proposed baseline constraint for uncalibrated cameras with overlap in **Section 3.2**.

### 3.1 Bundle Adjustment

Typical free-network BA minimizes the reprojection errors as noted in Eq (1):

$$\operatorname*{argmin}_{\mathbf{X},\mathbf{R},\mathbf{c},\mathbf{k}}(E_{reproj})$$
$$E_{reproj} = \frac{1}{2}\sum_{i,j} \rho\left(\left\|\mathbf{x}_{ij} - \pi\left(\mathbf{R}_j(\mathbf{X}_i - \mathbf{c}_j)\right)\right\|^2\right), \quad (1)$$

where $\mathbf{x}_{ij} \in \mathbb{R}^2$ is the position of the keypoint associated with the reconstructed 3D point $\mathbf{X}_i$ in camera $j$, $\mathbf{R}_j$ and $\mathbf{c}_j$ is the exterior orientation of camera $j$, $\mathbf{k}$ is the camera interior and distortion parameters, $\pi$ is the projection function, and $\rho$ is the robust Huber loss function (Hampel et al., 2011). We used the Levenberg Marquardt algorithm within Ceres Solver (Agarwal et al., 2020) to carry out all the optimizations presented in this paper.

### 3.2 Bundle Adjustment with Baseline Constraint for Uncalibrated Multi-Camera Systems with Overlap

As mentioned above, this case introduces our proposed solution for uncalibrated cameras that have overlap. In this case, the images from different video sequences are connected based on the overlapping views. Therefore, only one reconstruction result is generated and the relative orientation between images from different sequences can be obtained. The error function to be minimized is defined as:

$$\operatorname*{argmin}_{\mathbf{X},\mathbf{R},\mathbf{c},\mathbf{k}} \left(E_{reproj} + \lambda \frac{N_p}{N_t} E_{baseline}\right), \quad (2)$$

where $E_{baseline}$ is the proposed baseline constraint, $N_p$ is the number of reconstructed camera pairs, $N_t$ is the total number of pairs available in the dataset, and $\lambda$ is the weight parameter. Thus, the weight $\lambda \frac{N_p}{N_t}$ will change adaptively according to the size of the problem. The proposed baseline error term applies a weak constraint on the relative orientation between two cameras at different times. This baseline constraint applied for the camera network helps reduce errors that traditional BA cannot handle efficiently. In addition, formulating the error term in this way does not require the relative orientation of the cameras to be known while ensuring that they should be the same. The baseline error term is defined as:

$$E_{baseline} = \frac{1}{2}\sum_{i} \left(\|\mathbf{p}_i - \mathbf{p}_{i+1}\|^2\right), \quad (3)$$

where $\mathbf{p}_i \in \mathbb{R}^6$ denotes the relative orientation parameters between the two cameras at the time $i$. $\mathbf{p}_i$ contains three parameters of translation and three parameters of rotation. The rotation parameters are parameterized by axis-angle representation in the Euclidean space. $\lambda$ in Eq. (2) is empirically set to 500 during the reconstruction. Our experiments showed that a smaller $\lambda$ had less improvement of the drift problem, while a larger $\lambda$ led to a degenerate result. Once all cameras are reconstructed (i.e., $\frac{N_p}{N_t} = 1$ in Eq. (2)), we perform the last iteration that includes all the reconstructed cameras and 3D points. During this last iteration, we adaptively set $\lambda$ according to the value of the baseline error between two cameras as follows:

$$\lambda_i = \begin{cases} 250, & \left|(\mathbf{p}_i)_k - (\mathbf{p}_{avg})_k\right| \leq 5\left|(\mathbf{p}_{avg})_k\right| \\ 500, & \left|(\mathbf{p}_i)_k - (\mathbf{p}_{avg})_k\right| > 5\left|(\mathbf{p}_{avg})_k\right| \end{cases}, \quad (4)$$

where $\mathbf{p}_{avg}$ is the estimated average relative orientation using all the reconstructed pairs $\mathbf{p}_i$, $(\mathbf{p})_k$ denotes the k-th element of the vector $\mathbf{p}$. If any element of $\mathbf{p}_i$ satisfies the second condition in Eq (4), we regard the relative orientation of the camera pair at time $i$ as incorrectly estimated. Thus, we use $\mathbf{p}_{avg}$ to adaptively adjust $\lambda$ to penalize the pair of cameras that have a large error of the estimated relative orientation compared to $\mathbf{p}_{avg}$. In addition, this criterion is only applied after all cameras are reconstructed so that we can obtain a relatively accurate estimation of $\mathbf{p}_{avg}$.

## 4. EXPERIMENTAL RESULTS

Two experiments were performed to evaluate 1) the accuracy of 3D reconstruction considering different mounting configurations of cameras, and 2) the improvement of accuracy of 3D reconstruction using the proposed baseline constraint in BA for two uncalibrated cameras with overlap. Ground LiDAR point clouds were used as a reference for evaluation. We used OpenMVS (Cernea, 2020) to generate dense point clouds for all the experiments.

The data used in the experiments and the camera configurations are introduced in **Section 4.1**. The qualitative evaluation of each experiment is introduced in **Section 4.2**. The quantitative evaluation is introduced in **Section 4.3**.

### 4.1 Dataset

**Video frames.** Two GoPro Hero7 Black cameras were used to capture videos at a resolution of 2000 × 1500 with 30 fps. The cameras were mounted on the car, and we drove the car at an average speed of 25 mph on a closed-loop trajectory in a campus environment. We did not use special mounting frames for the cameras since our goal was to evaluate the proposed approach for uncalibrated cameras. Multiple mounting configurations were applied to collect data from different viewing angles: front, front-right, right, rear, and left. The cameras were coarsely synchronized based on the built-in timestamps when recording the videos. **Figure 1** shows an illustration of the mounting configurations of the cameras used in the experiments. **Table 1** shows the data size for each experiment. We extracted 33% from the captured video frames. Further, the images at the beginning of the video were matched with the images at the end. The collected data was in the same trajectory for all experiments conducted in this paper, but the number of extracted frames varied due to different traffic conditions.

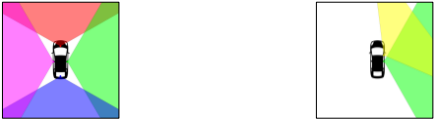

**Figure 1**. Mounting configurations of different cameras. Left: mounting configuration used for Experiment I. Right: mounting configuration used for Experiment II.

| Experiment | Camera configuration | # of images |
|---|---|---|
| I | Front (Red) | 2284 |
| | Right (Green) | 2047 |
| | Rear (Blue) | 2000 |
| | Left (Magenta) | 2047 |
| II | Right (Green) | 2000 |
| | Right (Yellow) | 2000 |

**Table 1**. The data size of each experiment.

**LiDAR data.** The ground LiDAR data was collected using Velodyne VLP-16 LiDAR sensors and geo-referenced using GNSS and IMU data. The mean accuracies were within 11 cm in the vertical direction and within 4 cm in the horizontal direction, and the standard deviations were between 1-5 cm for both directions. The sensor was installed on the car, which used the same trajectory for LiDAR data acquisition. **Figure 2** shows the top view of the LiDAR data, while **Table 2** shows the technical specification of the sensor.

### 4.2 Qualitative Evaluation

**Experiment I. Figure 3** shows the dense reconstruction results of four mounting configurations of cameras with different viewing angles. For the front and rear viewing angles, the reconstruction results contained both the left and right sides of the street scenes. Visual inspection showed that they both had drifted in horizontal and vertical directions, while the drift varied according to the facing directions of the cameras (i.e., principal ray). The reconstruction results of left and right cameras had less drift in the horizontal direction and more drift in the vertical direction. In contrast, the front and rear cameras showed less drift in the vertical direction but more drift in the horizontal direction. The visual inspection indicated that the drift was related to the viewing angles with regard to the moving direction of the car: the results tended to have more drift in the horizontal direction if the viewing angles of cameras were parallel to the moving direction of the car, and more drift in the vertical direction if the viewing angles of cameras were perpendicular to the moving direction of the car.

| Property | Value |
|---|---|
| Sensor model | Velodyne VLP-16 |
| Number of channels | 16 |
| Sampling frequency | 5-20 Hz |
| Angular resolution (horizontal/vertical) | 0.1°-0.4°/2.0° |
| FoV (horizontal/vertical) | 360°/30° |
| Range | 0.5-100 m |
| Range accuracy | Up to ±3 cm |

**Table 2**. Specifications of the LiDAR sensor used in evaluating the generated point clouds.

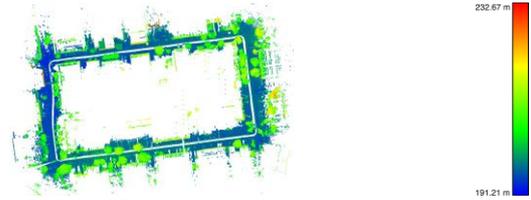

**Figure 2**. LiDAR data used for evaluation (color-coded by height values).

**Experiment II.** In this experiment, we evaluated the proposed baseline constraint in terms of reducing errors of reconstruction results for uncalibrated cameras with overlap (**Section 3.2**). **Figure 4** shows the dense reconstruction results of using traditional BA compared to the results using the baseline constraint in BA. The reconstruction results of using COLMAP with and without *rig_bundle_adjuster* were also used for comparison. We could observe that the results obtained using traditional BA had a drifted trajectory due to the accumulated errors. The region outlined in the green box (**Figure 4,** row 3) shows obvious drift in horizontal and vertical directions. The region outlined in purple and blue rectangles (**Figure 4,** row 3) further show both ends of the trajectory. Ideally, the two cameras should have the same relative orientation at different times since they were mounted on the car and kept static with each other during the data collection. However, due to the accumulation of errors, we could observe that several cameras had much bigger baselines than others because their corresponding cameras were localized to another end of the trajectory (purple and blue regions in **Figure 4**, row 3). On the other hand, the results obtained using the proposed baseline constraint in BA had a noticeable improvement, as shown in the yellow region (**Figure 4**, row 4). The baseline constraint enhanced the connection between the two uncalibrated cameras and prevented the two cameras from being

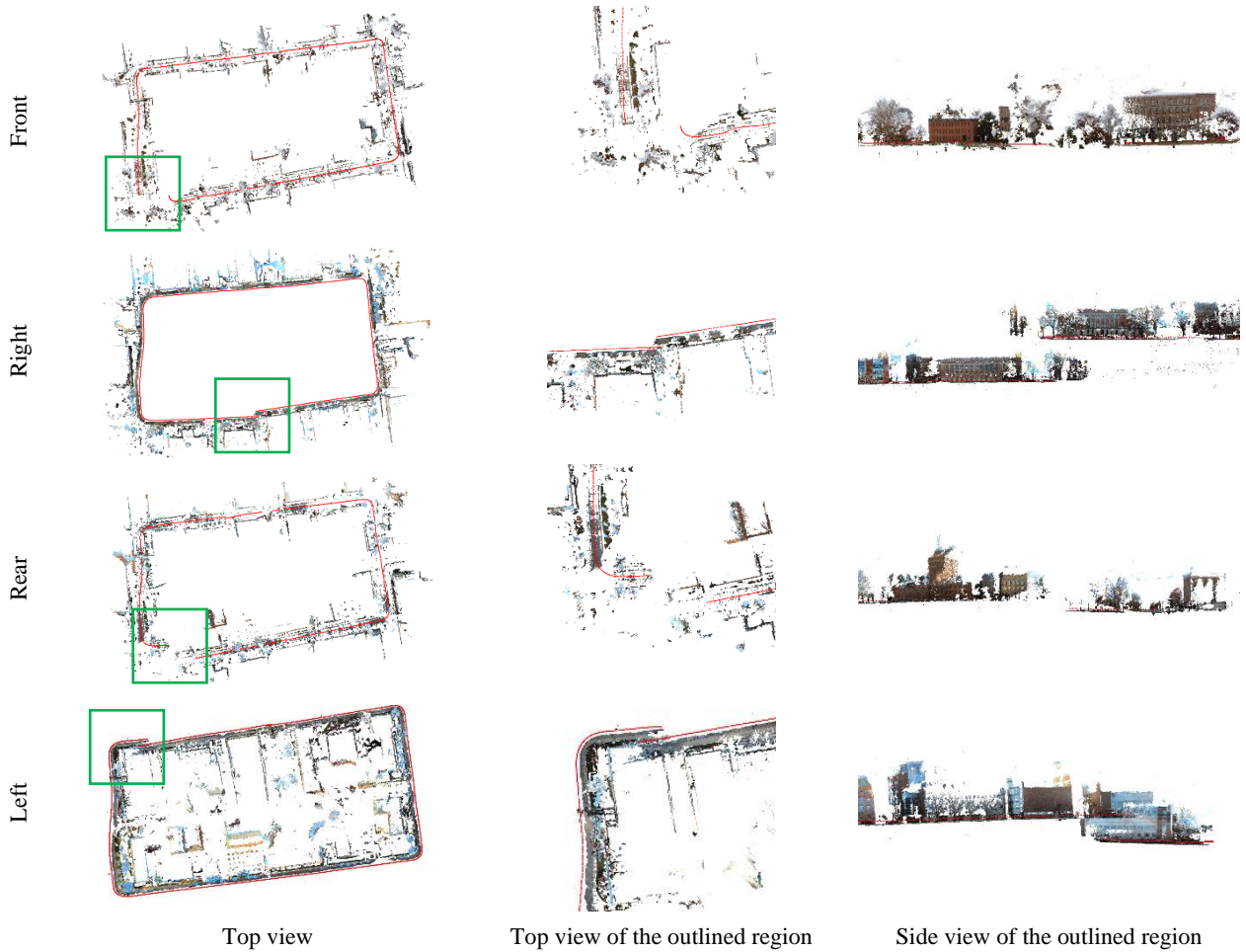

Top view      Top view of the outlined region      Side view of the outlined region deviated from each other. In addition, the baseline constraint significantly optimized the poses of cameras at both ends of the trajectory compared to traditional BA by moving them towards their corresponding cameras to reduce the absolute baseline errors; thus, the point clouds were also improved. On the other hand, *rig_bundle_adjuster* of COLMAP did not help reduce the drift problem, instead, it introduced more distortion which indicated that a worse result could be expected.

**Figure 3**. Visual results for dense reconstruction of four mounting configurations of cameras. The green rectangles outline the region with large drift. The estimated poses of cameras are shown in red.

### 4.3 Quantitative Evaluation

The quantitative evaluation was performed by measuring the mean absolute distance between the dense reconstruction results and the reference ground LiDAR point clouds using the open-source software CloudCompare (Girardeau-Montaut, 2015). For the front and rear cameras, the whole LiDAR point clouds were used since they covered both sides of the street scenes. For the other cameras covering only one side of the street, we segmented the LiDAR point clouds into left and right parts and computed the distance between each reconstruction with the corresponding segment of LiDAR point clouds. Since the generated reconstruction results were defined up to an arbitrary scale, we registered the dense point clouds to the LiDAR data before measuring the distance. In addition, we compared our results to the open-source library COLMAP (Schonberger and Frahm, 2016) in both experiments. In the second experiment, we used COLMAP's *rig_bundle_adjuster* which forced the relative orientation between both cameras at different times to be the same. For our method, we applied the proposed baseline constraint with the weights presented in **Section 3.2** in the BA.

**Experiment I. Table 3** shows the statistics of dense reconstruction accuracy of four mounting configurations of cameras in. In the results, the left camera showed the smallest mean absolute distance and standard deviation, which was consistent with the visual inspection in the qualitative evaluation. The results of the right camera had the largest mean absolute distance due to the largest drift in the vertical direction. The results of the front and rear cameras were very similar, which were also indicated by the similar dense reconstruction results in the qualitative evaluation.

| Camera configuration | Mean absolute distance [m] | | Std deviation [m] | |
|---|---|---|---|---|
| | Ours | COLMAP | Ours | COLMAP |
| Front | 1.253 | 2.593 | 2.049 | 3.920 |
| Right | 2.451 | 2.719 | 4.285 | 4.603 |
| Rear | 1.252 | 1.385 | 2.052 | 2.185 |
| Left | 0.916 | 1.645 | 1.653 | 2.733 |

**Table 3**. Statistics of dense reconstruction accuracy for different mounting configurations of monocular cameras (Experiment I).

**Experiment II.** In this experiment, we evaluated the results of using the constrained BA with our proposed baseline constraint compared to traditional BA. In addition, The reconstruction results of using COLMAP with and without *rig_bundle_adjuster* were also evaluated. **Table 4** shows the statistics of dense reconstruction accuracy. The mean absolute distance of using our proposed baseline constraint in BA was reduced from 2.028 to 1.432 meters, achieving an improvement of 29.38%. The standard deviation was also reduced from 2.965 to 2.105 meters with an improvement of 29.00%. In contrast, the results of using COLMAP's *rig_bundle_adjuster* were even worse than traditional BA, indicating the method used in COLMAP was not robust in this case. Compared to COLMAP's *rig_bundle_adjuster*, our proposed approach achieved an improvement of 74.24% in terms of mean absolute distance.

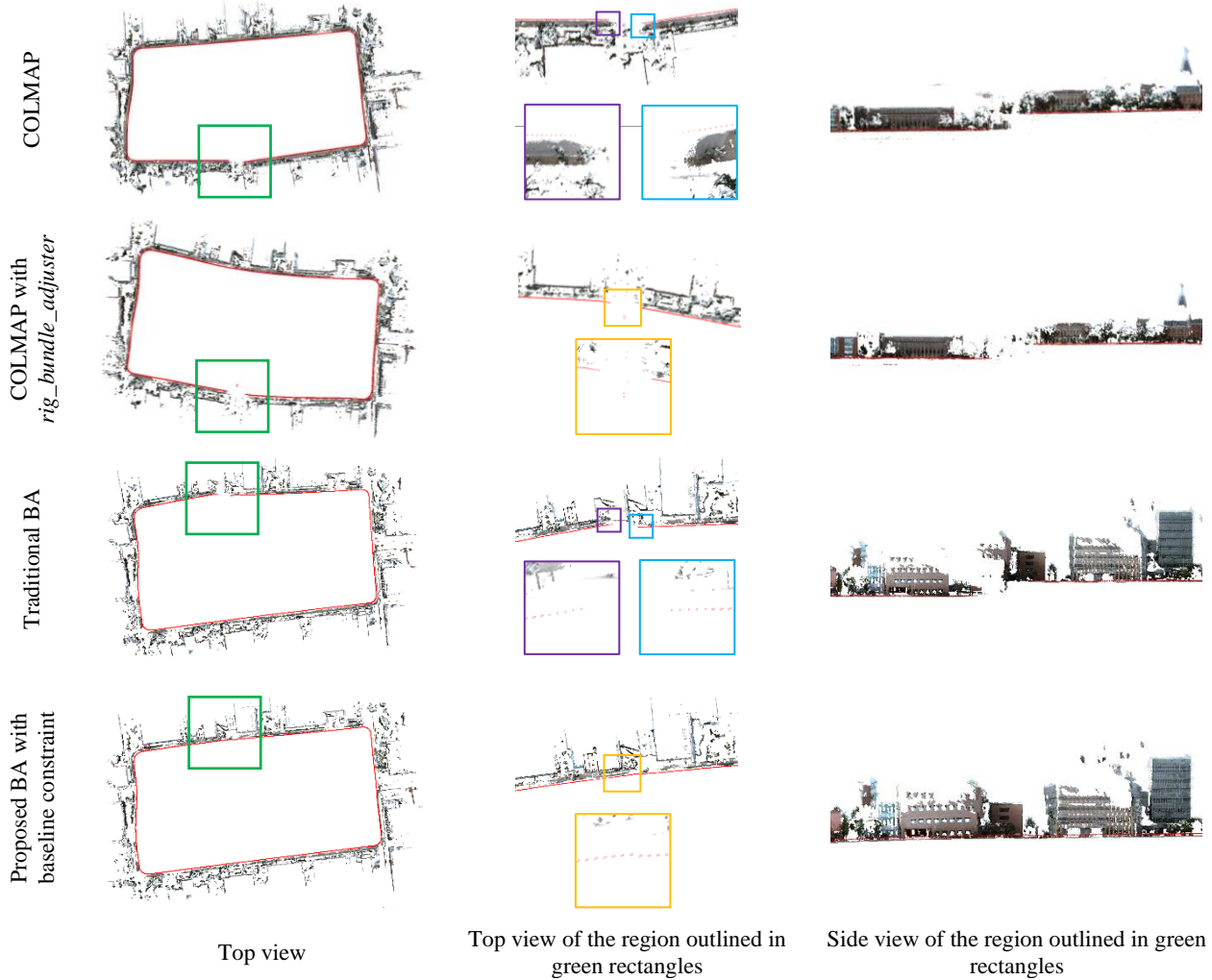

**Figure 4**. Visual results for dense reconstruction of two cameras with overlap. First row: reconstruction results of using COLMAP without *rig_bundle_adjuster*. Second row: reconstruction results of using COLMAP with *rig_bundle_adjuster*. Third row: reconstruction results of using our SfM pipeline with traditional BA. Fourth row: reconstruction results of using our SfM pipeline using the proposed baseline constraint. The green rectangles outline the regions with large drift (row 1-3) and the improvement (row 4). The purple and blue rectangles outline both ends of the drifted trajectory, where we could observe the poses of corresponding camera pairs were not correctly estimated. The orange rectangle (row 4) outlines the same region where our proposed baseline constraint improved the accuracy of the estimated camera poses, while COLMAP's *rig_bundle_adjuster* could not solve the drift problem. We could observe the impact of the baseline constraint (row 4) in correcting the drift in the estimated trajectory and producing more accurate point clouds compared to traditional BA (row 3)

| BA | Mean absolute distance [m] | | Std deviation [m] | |
|---|---|---|---|---|
| | Ours | COLMAP | Ours | COLMAP |
| Traditional | 2.028 | 2.976 | 2.965 | 3.690 |
| Constrained | **1.432** | 5.560 | **2.105** | 5.738 |

**Table 4**. Statistics of dense reconstruction accuracy for the uncalibrated multi-camera systems in overlapping case (Experiment II). COLMAP implements the constrained BA solution using the *rig_bundle_adjuster*.

## 5. CONCLUSION

In this paper, two experiments were performed to evaluate the accuracy of 3D reconstruction using different mounting configurations of cameras and the improvement of accuracy using our proposed baseline constraint in BA for the case in which two uncalibrated cameras have overlaps. The first experiment showed that different mounting configurations of cameras generated results with drift in both horizontal and vertical directions, while the degree of drift corresponded to the viewing angles of the cameras. The cameras with viewing angle parallel to the moving direction of the car tended to have more horizontal drift, and the cameras with viewing angle perpendicular to the moving direction of the car tended to have more vertical drift. The second experiment indicated that using our proposed baseline constraint in BA effectively improved the drift problems in the reconstruction results of two cameras with overlap. The baseline constraint restricted the camera poses from being deviated from the original trajectory and connected both ends of the open trajectory compared to the results using traditional BA. The statistical analysis showed an improvement of 29.38%, which was consistent with the qualitative evaluation. Our method was also compared with COLMAP's *rig_bundle_adjuster* and showed more accurate and robust results with 74.24% of improvement, as our method did not require the exact relative orientations between two cameras. The experiments showed that our robust minimization frameworks within BA could reduce errors and provide accurate 3D reconstruction by leveraging the baseline constraint between two cameras without knowing the calibration between them.

## ACKNOWLEDGEMENT

This work was supported in part by the Office of Naval Research (Award No. N000141712928 and N000142012141).

## REFERENCES

Agarwal, S., Mierle, K., Others, 2020. Ceres Solver.

Agarwal, S., Snavely, N., Seitz, S.M., Szeliski, R., 2010. Bundle adjustment in the large, European conference on computer vision. Springer, pp. 29-42.

Bay, H., Tuytelaars, T., Van Gool, L., 2006. SURF: Speeded Up Robust Features, in: Leonardis, A., Bischof, H., Pinz, A. (Eds.), European Conference on Computer Vision. Springer Berlin Heidelberg, Berlin, Heidelberg, pp. 404-417. doi.org/10.1007/11744023_32

Cavegn, S., Blaser, S., Nebiker, S., Haala, N., 2018. Robust and accurate image-based georeferencing exploiting relative orientation constraints. ISPRS Ann. Photogramm. Remote Sens. Spatial Inf. Sci. IV-2, 57-64. doi.org/10.5194/isprs-annals-IV-2-57-2018

Cernea, D., 2020. OpenMVS: Multi-View Stereo Reconstruction Library.

Esquivel, S., Koch, R., 2013. Structure from motion using rigidly coupled cameras without overlapping views, German Conference on Pattern Recognition. Springer, pp. 11-20. doi.org/10.1007/978-3-642-40602-7_2

Girardeau-Montaut, D., 2015. CloudCompare: 3D point cloud and mesh processing software.

Hampel, F.R., Ronchetti, E.M., Rousseeuw, P.J., Stahel, W.A., 2011. Robust statistics: the approach based on influence functions. John Wiley & Sons.

Häne, C., Heng, L., Lee, G.H., Fraundorfer, F., Furgale, P., Sattler, T., Pollefeys, M., 2017. 3D visual perception for self-driving cars using a multi-camera system: Calibration, mapping, localization, and obstacle detection. Image and Vision Computing 68, 14-27. doi.org/10.1016/J.IMAVIS.2017.07.003

Hartley, R., Zisserman, A., 2004. Multiple View Geometry in Computer Vision. Cambridge University Press. doi.org/10.1017/CBO9780511811685

Heng, L., Choi, B., Cui, Z., Geppert, M., Hu, S., Kuan, B., Liu, P., Nguyen, R., Yeo, Y.C., Geiger, A., 2019. Project autovision: Localization and 3d scene perception for an autonomous vehicle with a multi-camera system, International Conference on Robotics and Automation. IEEE, pp. 4695-4702. doi.org/10.1109/ICRA.2019.8793949

Heng, L., Furgale, P., Pollefeys, M., 2015a. Leveraging image‐based localization for infrastructure‐based calibration of a multi‐camera rig. Journal of Field Robotics 32, 775-802. doi.org/10.1002/rob.21540

Heng, L., Lee, G.H., Pollefeys, M., 2015b. Self-calibration and visual SLAM with a multi-camera system on a micro aerial vehicle. Autonomous robots 39, 259-277. doi.org/10.1007/s10514-015-9466-8

Lowe, D.G., 2004. Distinctive image features from scale-invariant keypoints. International Journal of Computer Vision 60, 91-110. doi.org/10.1023/B:VISI.0000029664.99615.94

Maset, E., Magri, L., Toschi, I., Fusiello, A., 2020. Bundle block adjustment with constrained relative orientations. ISPRS Ann. Photogramm. Remote Sens. Spatial Inf. Sci. V-2-2020, 49-55. doi.org/10.5194/isprs-annals-V-2-2020-49-2020

Maset, E., Rupnik, E., Pierrot-Deseilligny, M., Remondino, F., Fusiello, A., 2021. Exploiting Multi-Camera Constraints Within Bundle Block Adjustment: AN Experimental Comparison. The International Archives of Photogrammetry, Remote Sensing and Spatial Information Sciences 43, 33-38.

Muja, M., Lowe, D.G., 2014. Scalable nearest neighbor algorithms for high dimensional data. IEEE Transactions on Pattern Analysis and Machine Intelligence 36, 2227-2240. doi.org/10.1109/TPAMI.2014.2321376

Pierrot-Deseilligny, M., Jouin, D., Belvaux, J., Maillet, G., Girod, L., Rupnik, E., Muller, J., Daakir, M., Choqueux, G., Deveau, M., 2014. Micmac, apero, pastis and other beverages in a nutshell. Institut Géographique National.

Rublee, E., Rabaud, V., Konolige, K., Bradski, G., 2011. ORB: An efficient alternative to SIFT or SURF, International Conference on Computer Vision. doi.org/10.1109/ICCV.2011.6126544

Schonberger, J.L., Frahm, J.-M., 2016. Structure-from-motion revisited, IEEE/CVF Conference on Computer Vision and Pattern Recognition, pp. 4104-4113. doi.org/10.1109/CVPR.2016.445

Triggs, B., McLauchlan, P.F., Hartley, R.I., Fitzgibbon, A.W., 1999. Bundle adjustment—a modern synthesis, International workshop on vision algorithms, pp. 298-372. doi.org/10.1007/3-540-44480-7_21